\documentclass[letterpaper, 10 pt, conference]{ieeeconf}  % Comment this line out if you need a4paper

\IEEEoverridecommandlockouts                              % This command is only needed if 
\usepackage[utf8]{inputenc}
\usepackage{graphicx}
\usepackage{svg} % for including .svg via \includesvg
\usepackage{amsmath, amssymb}
\let\labelindent\relax
\usepackage{enumitem}
\usepackage{siunitx}
\usepackage{booktabs}
\usepackage{placeins} % for \FloatBarrier to control float placement
\usepackage{caption}
\usepackage{cite}
\usepackage[hidelinks]{hyperref}
\usepackage{tabularx}

\setlist{itemsep=4pt, topsep=4pt, leftmargin=*, labelsep=4pt}

\title{\LARGE \bf

A Factory-Floor Deployment Case Study of VLA Pipelines for Industrial Packaging Task: Workflow, Failures, and Lessons
}

\author{Siemens Corporation$^{1}$%
 \thanks{$^{1}$See Acknowledgments for author list.}}

\begin{document}

\maketitle
\thispagestyle{empty}
\pagestyle{empty}

%%%%%%%%%%%%%%%%%%%%%%%%%%%%%%%%%%%%%%%%%%%%%%%%%%%%%%%%%%%%%%%%%%%%%%%%%%%%%%%%
\begin{abstract}

Vision-Language-Action (VLA) policies have shown promising manipulation capabilities, yet their practical impact is often limited by the reliability demands of real-world deployment. 
We present a deployment study of an industrial packaging task at Siemens Factory (GWE, Erlangen, Germany), where a robot must pick a transparent accessory bag from a cluttered pile, insert it into the remaining cavity of a cardboard package, and ensure that the bag and its contents remain below the closing plane. 
Our goal is to understand the practical effort required to adapt a pretrained Pi0.5 policy to a single factory-floor task through iterative fine-tuning and deployment-driven refinement. 
The pipeline consists of repeated loops of data collection, curation, fine-tuning, evaluation, and targeted recovery data collection. 
We have accumulated 2535 episodes (\(\sim\)10 hours) from the on-site factory settings. 
In this paper, we contribute an empirical account of a factory-floor VLA deployment, highlighting recurring failure modes and  lessons that inform how to improve the deployment workflow.

\end{abstract}

%%%%%%%%%%%%%%%%%%%%%%%%%%%%%%%%%%%%%%%%%%%%%%%%%%%%%%%%%%%%%%%%%%%%%%%%%%%%%%%%

\section{Introduction}

Vision-Language-Action (VLA) models~\cite{kim2024openvlaopensourcevisionlanguageactionmodel,nvidia2025gr00tn1openfoundation,black2026pi0visionlanguageactionflowmodel,intelligence2025pi05visionlanguageactionmodelopenworld} have recently emerged as promising general-purpose robot policies, with impressive demonstrations across a wide range of manipulation tasks.
However, many of these results are reported in controlled lab settings, where scenes, objects, and success criteria are more forgiving than in production environments.
In contrast, factory-floor deployment imposes stringent operational constraints---throughput, robustness to occlusion, low latency, safe execution, and tight downstream quality requirements, that can expose brittle behaviors even when lab performance looks strong.
As a result, despite rapid progress in foundation models for robotics, the gap between successful lab demonstrations and production-ready behavior in industrial settings remains significant.

Recent efforts have begun to study this gap more directly. Wang et al.~\cite{pi0-experiment-wild} evaluate $\pi_{0}$ in the wild over more than 300 zero-shot real-world manipulation trials, revealing both promising general-purpose capabilities and important practical failure modes.
Likewise, LeRobot’s shirt-folding system~\cite{kooijmans2026_unfolding_robotics_the_open_source_recipe_for_teaching_a_robot_to_fold_your_clothes} demonstrates that high performance on a challenging deformable-manipulation task is possible when the entire pipeline from data collection to deployment is carefully engineered.
While these results are encouraging, they do not fully address the realities of industrial production tasks, where integration with upstream and downstream processes, safety practices, and runtime constraints can make dependable operation substantially more demanding.

\begin{figure}[t]
  \centering
  \includegraphics[width=\columnwidth]{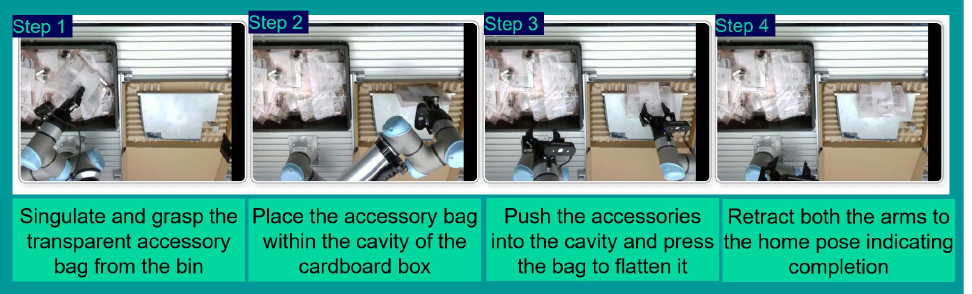}
  \caption{Overview of the seasoning task and the steps to complete the task.}
  \label{fig:seasoning-task}
\end{figure}

This gap motivates the present study. Rather than asking whether a foundation model can generalize across many tasks, we focus on a narrower but practically important question: what does it take to adapt a pretrained VLA model for a single industrial task, and what limits its reliability during real deployment?
In particular, we consider only task-specific fine-tuning and do not address pretraining.
Our goal is to characterize the concrete effort required to narrow the deployment gap on a factory floor, including how much task data is needed for fine-tuning, how data should be collected iteratively to address observed failures, and what additional challenges arise during actual deployment.

We study these questions through a real factory packaging use-case.
The goal is to pick a transparent accessory bag containing a user manual and an industrial connector cable from a bin with a pile of similar bags and insert it into the available cavity of a cardboard package that already contains the primary product as shown in Figure~\ref{fig:seasoning-task}.
While the task is simple to describe, meeting production requirements is challenging in practice.
Key difficulties include singulating a transparent bag from clutter, handling variation in grasp points arising from different bag orientations, and ensuring that the contents of the bag settle inside the package cavity rather than on top of the packed product.
In addition, the final packed state must satisfy a strict downstream requirement: the accessory bag and its contents must not protrude above the product surface, since the package must close flat after insertion.
Together, these task-specific constraints make the packaging task a useful case study for examining the practical bottlenecks that arise when fine-tuning and deploying a VLA policy on the factory floor.

Building on this task setting, we report a real factory-floor deployment study on adapting a pretrained VLA policy to an industrial packaging task.
Our contributions are:
\begin{enumerate}
   \item \textbf{Factory-floor deployment study and workflow:} an end-to-end account of adapting a pretrained VLA to an industrial packaging task, detailing the iterative onsite workflow (rollouts, failure analysis, targeted data collection, curation, and fine-tuning) and the practical considerations that drove key design decisions.
    \item \textbf{Failure taxonomy:} a qualitative and quantitative analysis of common failure modes observed during factory-floor evaluation.
    \item \textbf{Lessons learned from industrial VLA deployment:}  deployment-driven takeaways that generalize beyond our setup, covering teleoperation, data curation and runtime engineering.
\end{enumerate}

\section{Hardware \& Software}

For our experiments, we used a pair of UR7e robots equipped with Robotiq 2F-85 grippers. Custom 3D-printed fingertips were installed on the gripper to make it easier to manipulate the plastic bags.
Additional light panels were installed to ensure controlled lighting conditions in the work-cell.
To further reduce the environmental variance, we installed additional fixtures to keep the cardboard box and bin in a fixed position.

For data collection, we used the tracking from a Meta Quest 3 to control the end-effector pose of the robots. A significant amount of effort was spent on refining the teleoperation stack such that it was smooth, responsive, and safe. This included the implementation of features such as interpolation of lower frequency commands to high frequency commands that the robots expect (500hz for URs), impedance control for compliance, and motion planning with collision avoidance for safety.
Image observations were captured through the wrist-mounted cameras on the robot arms and an additional base camera mounted on the work-cell.

All rollouts were run on a Siemens Industrial PC (IPC) equipped with an RTX 5090 GPU. We also implemented asynchronous inference with real-time chunking \cite{black2025realtimeexecutionactionchunking}, which we found was critical for ensuring the trained policies run smoothly and accurately.

\section{Execution Strategy}\label{sec:exec}
The deployment process was split into two major phases.
The first phase was conducted at a mock robot cell in one of Siemens’ internal research labs.
This was used to validate the hardware stack and the data collection and training workflow.
Over the rounds of data collection, training, and evaluation, we refined an execution strategy that was consistent to teleoperate.
In particular, the following steps that human typically would do were hard to execute on the robot and/or reproduce with VLAs:
\begin{itemize}
    \item Shaking the bag to have the contents of the bag settle to the bottom. The repetitive nature of the motion is challenging for VLAs to understand and replicate, as they lack memory by design. In addition, the motion is typically executed with high-speed and was challenging to teleoperate consistently.
    \item Curling the bag to fit in the cavity of the package. The motion required a large amount of dexterity and was hard to teleoperate consistently.
\end{itemize}

This prompted us to make the following adjustments to final execution strategy, which was the following:
\begin{enumerate}
    \item With one arm, grasp a bag from the pile in the bin, ensuring that the contents of the bag were not within the gripper.
    \item The arm then lifts the bag, using gravity to help let the parts settle to the bottom of the bag before transporting it to the package where it inserts the bag into the cavity.
    \item The other arm then pushes any protruding accessories into the cavity and then presses the bag onto the product in the package to ensure that when the box is closed, the bag is flattened.
\end{enumerate}

In this process, around 900 episodes were collected. 
We found that once we developed an execution strategy that was consistent to teleoperate, the speed of data collection greatly increased and there were significantly fewer errors across episodes. 
This is critical for the next phase, where a similar robot cell was deployed in the factory floor followed by additional rounds of data collection and training.

\section{Data Collection and Training}

\begin{figure}[t] \centering \includegraphics[width=\linewidth]{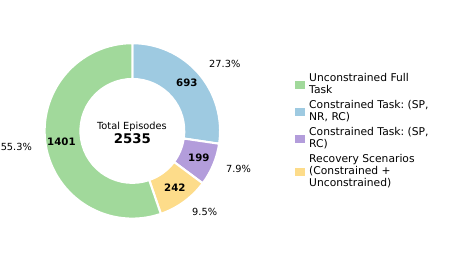} \caption{Breakdown of the dataset by task scenario category. Early data collection emphasized constrained scenarios for validating core behaviors, recovery scenarios, and finally unconstrained full-task scenarios. Inner labels denote episode counts and outer labels denote percentages of the total dataset (2,535 episodes). Constraints applied: Settled Parts (SP), No Reposition (NR), Reduced Count (RC).} \label{fig:data-collection-chart} \end{figure}

The second phase focused on onsite data collection and deployment-driven refinement on the factory floor under a limited deployment window of approximately two weeks. 
Rather than collecting a single dataset under full task complexity, we adopted an iterative strategy that progressively increased scenario difficulty.
The goal was to validate the main behaviors required by the execution strategy in Section~\ref{sec:exec} before attempting the full production-like setting.

In particular, these were the behaviors we wanted to validate:
\begin{enumerate}
    \item Perceiving the edges of bags in dense piles and identifying “empty” portions of the bag as grasping points.
    \item Identifying if the bag is not possible to grasp safely (e.g., colliding with the bin walls) and if so repositioning the bag with additional manipulations.
    \item After the bag is placed, verifying if some contents of the bag are outside the cavity and push them back in.
\end{enumerate}

To validate these behaviors efficiently within the limited onsite window, we introduced a set of environmental constraints and execute successive rounds data collection, training, and evaluation, removing the constraints one-by-one after each round. The constraints are the following:

\textit{Settled Parts (SP).} Contents of the bag are on one side of the bag, creating a distinguishable “empty” and “filled” side.

\textit{No Reposition (NR).} Bags are placed in the bin such that they can be safely grasped without repositioning them.

\textit{Reduced Count (RC).} Number of bags in the bin are reduced to make it easy to distinguish the edges of the bags.

We conducted three rounds of data collection. The first round (693 episodes) included all constraints to simplify the task through reduced clutter and favorable bag configurations to validate the nominal execution strategy. The second round (199 episodes) removed the \textit{No Reposition} constraint to validate that the policy can execute corrective motions to re-position the bags and then grasp them to avoid collision with the bin or grasping the contents of the bag. The third round of data collection (1401 episodes) removed all constraints to match the production setting, which included denser piles and fully scrambled bags. The resulting dataset distribution is summarized in Figure~\ref{fig:data-collection-chart}.

After each round of data collection, all episodes were manually reviewed and erroneous trajectories (less than 5\% of the data) were removed. All data collected so far was then used to fine-tune the base checkpoint of $\pi_{0.5}$ via OpenPI. For the first and second round of data collection, we used LoRA fine-tuning with batch size 32 for 30k steps, as the primary goal was to validate specific policy behaviors and the memory footprint was small enough for the RTX 5090 GPU in the IPC. After the third round of data collection, we switched to full fine-tuning with batch size 128 for 60k steps (\(\sim\)4 epochs) on Siemens' internal compute cluster.

Between each round, we reviewed evaluation results and collected \textit{Recovery} data (242 episodes), covering scenarios such as failed grasps, multi-bag picks, or imperfect placement outcomes. 
In addition, we also required that the policy achieve at least 70\% success during evaluation before proceeding to the next round of data collection.
This iterative process allowed us to make the most of our limited time on-site, as each round of training and evaluation helped us identify successful behaviors and failure modes. This diagnostic information was then used to update the data collection strategy (e.g., collecting recovery data targeting a particular failure state).

\begin{figure}[!t]
    \centering
    \includegraphics[width=\linewidth]{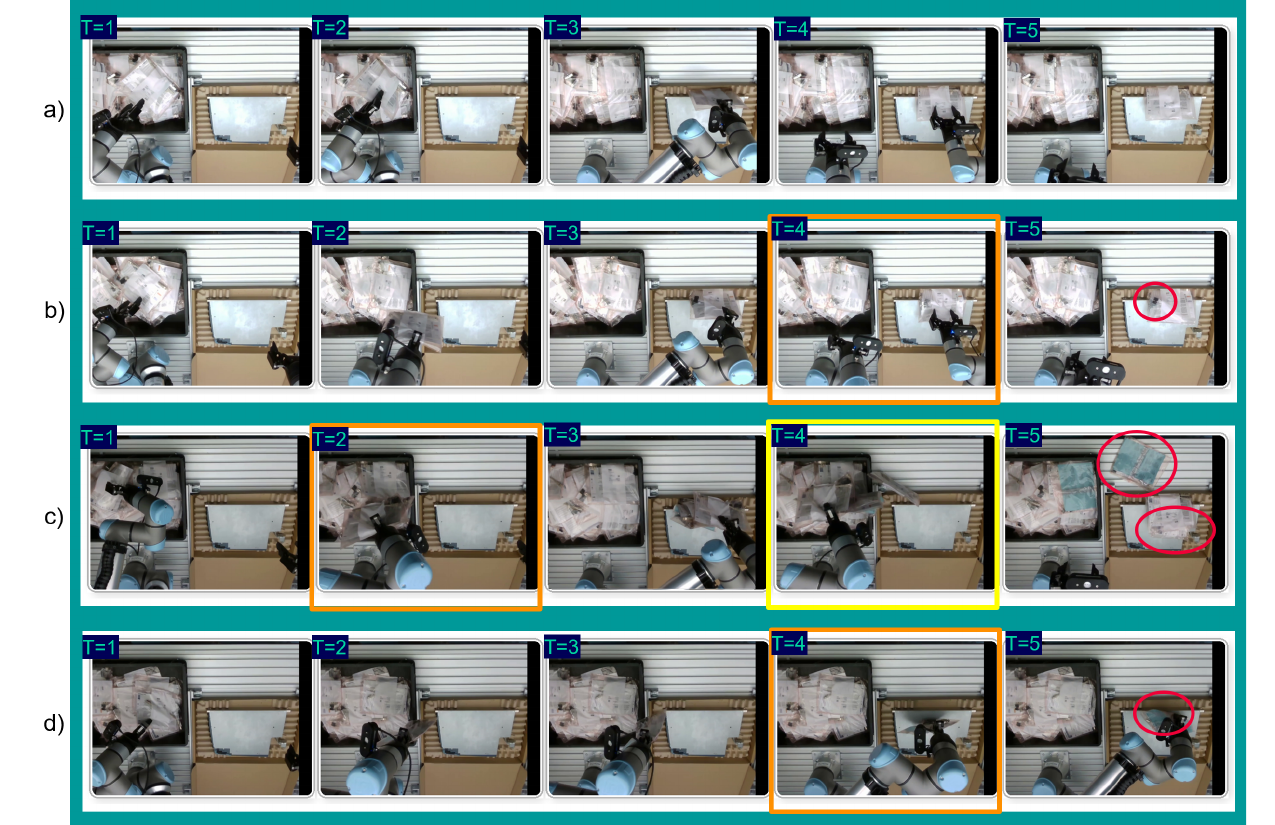}
    % \begingroup
    % \setlength{\abovecaptionskip}{2pt}
    % \setlength{\belowcaptionskip}{0pt}
    \caption{Successive rollout frames from onsite evaluations. Section (a) shows a successful rollout. Section (b)--(d) show the most common failure modes: (b) bag contents remain on top of the product, (c) multiple bags are grasped and bag dropped outside the box, (d) the bag is not fully inserted into the package box. For each failure case, the orange boxed frame marks when the failure first occurs, the red circle in the final frame highlights the resulting error state, and the yellow boxed frame (when present) indicates a recovery attempt that still results in failure.}
    % \endgroup
    \label{fig:three-common-failures}
\end{figure}

\section{Results}

After the last round of data collection and training, we ran a more comprehensive round of evaluations. We ran multiple trials with different scenarios, where we simulate the policy emptying a bin with a pile of 30 randomly placed bags. Trial 1 included the \textit{Settled Parts} constraint while Trials 2 and 3 were fully unconstrained. For each episode, the policy attempts to complete the task for a single bag (i.e., one pick-and-place operation). Episodes are given a time limit of one minute. Regardless of success or failure, the bag the policy is attempting to transport is removed at the end of the episode, so at the end of each trial the bin will be empty.

\renewcommand{\dbltopfraction}{0.9}
\renewcommand{\dblfloatpagefraction}{0.8}
\begin{table*}[!tp]
    \small
    \centering
    \caption{Categories of Failures and Their Frequencies for Trials 2 \& 3}
    \label{tab:failure analysis}
    % \footnotesize % Removed from here
    % \setlength{\tabcolsep}{2.5pt} % Reduced column separation
    % \renewcommand{\arraystretch}{1.1}
    % \resizebox{\textwidth}{!}{% <--- IMPORTANT CHANGE: \textwidth instead of \columnwidth
    \begin{tabular*}{\linewidth}{@{\extracolsep{\fill}}lccc@{}} % Failed-episode columns only
        \toprule
        & \textbf{Trial 2} & \textbf{Trial 3} & \textbf{Overall} \\
        \cmidrule(lr){2-2} \cmidrule(lr){3-3} \cmidrule(lr){4-4}
        \textbf{Failure Type} & \textbf{\%} & \textbf{\%} & \textbf{\%} \\
        \midrule
        Bag Contents Remain on Product & $62\%$ & $69\%$ & $65\%$ \\
        Multiple Bags Grasped & $23\%$ & $23\%$ & $23\%$ \\
        Bag Not Fully Inserted into Box & $23\%$ & $7.7\%$ & $15\%$ \\
        Poor or Failed Grasps of Bag & $23\%$ & $7.7\%$ & $15\%$ \\
        \bottomrule
    \end{tabular*}%
\end{table*}

For each failed episode in Trials 2 and 3, we recorded a short description of the errors the policy made over the course of the episode. These descriptions were then categorized into a set of common mistakes, which are visualized in Figure \ref{fig:three-common-failures} and quantified in Table \ref{tab:failure analysis}. Note that some episodes may contain more than one error. The most common mistakes were the following:

\textbf{Bag Contents Remain on Product (65\% of failed episodes).} This can happen for multiple reasons. The policy can fail to identify accessories lying on top of the product if they are small or hidden under the instruction manual. Challenging dynamics prevents the policy from pushing the parts into the cavity (e.g., we have observed the hook of the ethernet cable clipping the instruction manual).

\textbf{Multiple Bags Grasped (23\% of failed episodes).} The policy fails to correctly distinguish the borders between the transparent bags and ends up transporting multiple bags to the box.

\textbf{Bag Not Fully Inserted into Box (15\% of failed episodes).} This can happen for multiple reasons. The policy can fail to perceive the correct location to place the bag, as the view from the wrist camera is highly occluded and at an extreme angle. The policy can also grasp the bag from a poor angle, and the error cascades into a failed insertion into the box.

\textbf{Poor or Failed Grasps of the Bag (15\% of failed episodes).} The policy either: (1) does not scoop deep enough to grasp a bag, (2) grasps from a poor angle that causes the bag to be unstable during transport, or (3) fails to identify an empty region of the bag to grasp.

\section{Discussion}

While the final success rates of the finetuned policy did not meet expectations, there were many valuable lessons that were learned over the course of the deployment process, including the following:

\textbf{Spending the engineering effort to ensure that the control stack is responsive} is critical for high-quality data collection and performant policies. Latency and jitter distort the teleoperator's perception--action loop: by the time feedback arrives, the scene may have already evolved, encouraging over-corrections and yielding demonstrations that are less smooth and less representative of the intended behavior. Implementing asynchronous inference loops (e.g., with real-time chunking) increases policy throughput and precision.

\textbf{Data collection, training, and evaluation needs to be done in an iterative loop}, which allows you to (1) refine the execution strategy to make data collection efficient and consistent, (2) empirically validate policy behavior with focused experiments, and (3) identify adjustments to the data collection process such as the integration of recovery behaviors.

\textbf{Detailed cataloging and reviewing of data} are critical for reducing errors in the dataset and understanding how segments of the data contribute to policy performance.

\textbf{Qualitative analysis during evaluation provides invaluable information on top of binary success rates}. In particular, the categorization of failure modes provides quantifiable metrics that lead to actionable improvements.

In the future, we feel that the following adjustments would greatly boost the success of VLA deployments:

\textbf{Spending more time to reduce the morphology gap in the hardware setup.} For example, adjusting the robot mount such that the end-effector is in a more ergonomic pose can potentially enable the teleoperator to execute motions that simplify the execution strategy but were awkward or impossible to reproduce. 

\textbf{Implementing a human-in-the-loop workflow during evaluations} (e.g., interrupting the execution and handing control to a teleoperator) would make collection of recovery behaviors more effective and efficient. Currently, we first teleoperate the robot to a failure state instead of letting the policy do so, which may not match the distribution of actual failure states.

\textbf{Further validating that the inputs in the data provide enough information.} We found that a large proportion of failures originated from the cameras having a poor view of the objects of interest. Making minor adjustments to the execution strategy, such as utilizing the wrist camera of the second arm to provide an additional view, can expand what the policy is observing.

\textbf{Integrating memory into the VLA} may enable to the policy to reproduce a larger set of dexterous behaviors, reducing the number of compromises made when developing the execution strategy. For example, we had to remove repetitive behaviors such as shaking, which is useful for helping accessories settle to the bottom of the bag.

\textbf{Implementing more granular metrics} such as sub-task success rate and progress completion on top of qualitative analysis would add quantifiable metrics beyond binary success rate that make easier to prove if a policy is better or not. With longer evaluation processes, it becomes important to speed up the evaluation process with automations, as the process is currently manual and time-consuming.

\section{Conclusion}

We presented a deployment study of the application of VLAs for an industrial packaging task, which involves the placement of an accessory bag within the remaining cavity in a package. We detail the steps we took to refine our hardware and software setup, execution strategy, and data collection and training workflow. We then presented the overall results from our on-site deployment along with a qualitative analysis of the failures. We then share the steps in the methodology that we believe are critical for future deployments, as well as adjustments that would improve the potential for VLA deployments for industrial tasks.

\addtolength{\textheight}{-12cm}   % This command serves to balance the column lengths

%%%%%%%%%%%%%%%%%%%%%%%%%%%%%%%%%%%%%%%%%%%%%%%%%%%%%%%%%%%%%%%%%%%%%%%%%%%%%%%%

%%%%%%%%%%%%%%%%%%%%%%%%%%%%%%%%%%%%%%%%%%%%%%%%%%%%%%%%%%%%%%%%%%%%%%%%%%%%%%%%

%%%%%%%%%%%%%%%%%%%%%%%%%%%%%%%%%%%%%%%%%%%%%%%%%%%%%%%%%%%%%%%%%%%%%%%%%%%%%%%%
\newpage

\section*{Acknowledgments}
\noindent Project Members: Brian Zhu, Philipp Schmitt, Philine Meister, Lukas Gensler, Momen Khalil, Emmanuele Poggi, Johannes Hechtl, Carsten Braunroth, Kai Wurm, Gokul Narayanan, Eugen Solowjow, Georg von Wichert, Andre Scholz, Felix Albrecht, Maxmillian Metzner

%%%%%%%%%%%%%%%%%%%%%%%%%%%%%%%%%%%%%%%%%%%%%%%%%%%%%%%%%%%%%%%%%%%%%%%%%%%%%%%%

\bibliographystyle{unsrt}
\bibliography{refs}

\end{document}